\documentclass[runningheads]{eccv2020kit/llncs}
\usepackage{graphicx}
\usepackage{comment}
\usepackage{amsmath,amssymb} 
\usepackage{color}

\usepackage{enumitem}

\usepackage{xcolor,soul}
\sethlcolor{lightgray}
\usepackage{multicol}
\usepackage{hyperref}
\usepackage{caption}
\usepackage{subcaption}

\usepackage{xcolor}

\usepackage{float}

\usepackage[export]{adjustbox}
\usepackage{booktabs}
\graphicspath{ {images/} }
\newcommand{\R}{\mathbb{R}}

\begin{document}
\pagestyle{headings}
\mainmatter
\def\ECCVSubNumber{}  

\title{Demographic Influences on Contemporary Art with Unsupervised Style Embeddings} 

\titlerunning{Demographic Influences on Contemporary Art}
%
\author{Nikolai Huckle\inst{1} \and
Noa Garcia\inst{2} \and
Yuta Nakashima\inst{2}}
\authorrunning{Huckle et al.}
%
\institute{University of Bamberg \email{n.huckle@posteo.de} \and
Osaka University
\email{\{noagarcia,n-yuta\}@ids.osaka-u.ac.jp}}
\maketitle

\begin{abstract}
Computational art analysis has, through its reliance on classification tasks, prioritised historical datasets in which the artworks are already well sorted with the necessary annotations. Art produced today, on the other hand, is numerous and easily accessible, through the internet and social networks that are used by professional and amateur artists alike to display their work. Although this art---yet unsorted in terms of style and genre---is less suited for supervised analysis, the data sources come with novel information that may help frame the visual content in equally novel ways. As a first step in this direction, we present contempArt, a multi-modal dataset of exclusively contemporary artworks. contempArt is a collection of paintings and drawings, a detailed graph network based on social connections on Instagram and additional socio-demographic information; all attached to 442 artists at the beginning of their career. We evaluate three methods suited for generating unsupervised style embeddings of images and correlate them with the remaining data. We find no connections between visual style on the one hand and social proximity, gender, and nationality on the other.
\end{abstract}

\section{Introduction}

The methodological melting pot that is the interdisciplinary field of digital art history has, in recent years, been shaped more by exhausting technical novelty than theoretical guidance~\cite{bishop2018against,badea2018can}. Alongside proponents of the technological transformation, who see its vast databases as an opportunity to investigate large-scale patterns of various nature~\cite{drucker2013there,jaskot2019digital}, there has been criticism that its pure empiricism hinders any true discovery~\cite{bishop2018against}. Computer vision (CV), which has also found its way into the field by providing state-of-the-art methods~\cite{karayev2013recognizing} and assembling large multi-modal datasets~\cite{mao2017deepart,bianco2019multitask,garcia2018read}, has not been exempt from this criticism. Specifically, that only extracting and connecting high-level semantics of paintings ignores the real-world context in which art is being produced and belongs to an outdated form of comparative art history~\cite{mercuriali2019digital,abe2017introduction}.

Furthermore, recent progress, both visually~\cite{zhu2017unpaired,jing2019neural} and numerically~\cite{bianco2019multitask}, has not changed the fact that CV's potential to effectively engage the digital humanities is bounded by one recurrent factor: labels. Labels are an obvious necessity for aligning input data with a relevant supervisory signal, in general learning tasks, and a less obvious one in creating image tuples for texture synthesis or generative models. As classification tasks have become omni-present throughout the field, so have labels. At first glance, giving centre stage to typology seems to be in line with it being one of art historians main research interests~\cite{jaskot2020digital}. 

However, in supervised learning, the annotations serve as research’s means and not its end, rendering the possibility of expanding upon that same annotation impossible. This becomes problematic due to the absence of perfect knowledge in art history, as opposed to more common classification tasks such as object recognition, where the classes are flawless and the image labels non-negotiable~\cite{lin2014microsoft,xiao2017fashion}. Contrary to images of objects, paintings and their historical contextualisation is very much an open-ended and contested subject~\cite{jaskot2019digital}. By ignoring the uncertainty attached to the labels of art datasets, CV on the one hand handicaps its potential in investigating art in a novel way and, on the other hand perpetuates a misleadingly homogeneous image of the art-historical canon.

Overcoming these limitations and advancing into interdisciplinary depths requires CV to turn away from existing labels and instead embrace two other central research interests in classical art history: a) the visual representation of art and b) the social context of its production~\cite{jaskot2020digital}. In this work, we present two contributions in line with these two themes:
\begin{enumerate}
    \item For extracting visual representations in an unsupervised manner, we evaluate and utilise existing techniques from different domains.
    \item For studying the social world surrounding art, we introduce contempArt, the first dataset on contemporary painters with images, socio-demographic information and social media data representing social relationships.
\end{enumerate}

Aligning the information on demographics and social relationships with the attained style-embeddings allows us to investigate tangible connections beyond the visual realm. However, we find no evidence that social closeness entails similarity in style or that demographic factors correlate with visual content. 
\section{Related Work}

\subsection{Unsupervised Art Analysis}
\paragraph{Analysis of embeddings.} Compared to the substantial amount of work on art classification~\cite{shamir2010impressionism,mensink2014rijksmuseum,mao2017deepart,strezoski2018omniart,bianco2019multitask}, only rarely have image representations themselves been at the centre of computational analysis. One of the earliest such works is the seminal paper by~\cite{taylor1999fractal}, in which the fractal dimension of Jackson Pollock’s drip paintings is measured and compared over the course of his career. In a similar vein, aggregate image-level statistics based on colour and brightness are used in~\cite{sigaki2018history,kim2014large,lee2018heterogeneity} to visualise the evolution of artworks. In~\cite{elgammal2015quantifying}, object recognition features and time annotations are combined to assign creativity scores representing the visual novelty of artworks, at the time of their creation. Due to the success of convolutional neural networks (CNN) in image classification, art analysis has seen handcrafted image representations being replaced by deep feature vectors extracted from these CNNs. Of note is~\cite{karayev2013recognizing}, in which deep features from a CNN trained for object recognition outperform older handcrafted features in classifying art styles.~\cite{cetinic2019deep} also uses features provided by multiple CNNs to investigate connections between artwork and human perception.~\cite{brachmann2017using}, on the other hand, analyses different variance statistics between the layers of an object-recognition CNN and finds that these values can discern art from non-art.

\paragraph{Analysis of Clusters.} Other work is focused on applying complex unsupervised learning techniques on both handcrafted and deep image features~\cite{spehr2009image,shamir2012computer,gultepe2018predicting,castellano2020deep}. Most notable amongst these clustering studies is~\cite{wynen2018unsupervised}, in which artistic style is attained by computing statistics at different layers of a pre-trained object recognition CNN, a methodology created for texture synthesis \cite{gatys2016image}, and these features are additionally clustered with archetypal analysis~\cite{cutler1994archetypal}.

\subsection{Social Context}

Expanding art-based deep learning techniques to include information beyond the visual has been premiered in~\cite{garcia2019context}, where multi-modal relationships between various artistic metadata are used to increase the performance of image classification and image retrieval. To allow these broader analysis, the Semart dataset~\cite{garcia2018read} was introduced, where images are paired with artistic comments describing them. 

An older dataset on deviantArt,\footnote{\url{https://www.deviantart.com/}} one of the largest online social networks for art, initially only contained information on measures of social connections~\cite{salah2012deviantart} but was later expanded to include large amounts of image data~\cite{salah2013combining}. Although it has been used both for simple social network~\cite{salah2012deviantart,salah2013combining} and image analysis~\cite{yazdani2017quantifying} never have these data sources been combined in a scientific analysis.~\cite{kim2017creative}, exploring a dataset on another creative online community named Behance,\footnote{\url{https://www.behance.net/}} does combine artistic images and social network relationships but the visual information is aggregated too coarsely as to have any meaning.

Among the more commonly used art datasets such as Wikiart\footnote{\url{https://www.wikiart.org/}} or~Painting91 \cite{khan2014painting}, deviantArt especially stands out, as it is focused exclusively on contemporary art and, more importantly, work that is being produced outside of the commercialised world of galleries, auction houses and museums. The disadvantage being that prospective artists cannot be separated from amateurs, making their joint study more related to cultural analytics than digital art history. As detailed biographical data is never mandatory on online social networks, their user-generated image content can be easily transformed into scientific datasets, but important annotations, characterising and validating the creators of said content, are lost or never available at the start. With contempArt, we reduce the conventional scope by only including students enrolled at German art schools, thereby guaranteeing detailed socio-demographic information but sacrificing quantity. The result is a unique but narrow snapshot of contemporary painting culture.
\section{contempArt Dataset}
\begin{figure}[t]
    \centering
    \begin{subfigure}[t]{.24\textwidth}
        \centering\includegraphics[width=1.0\linewidth]{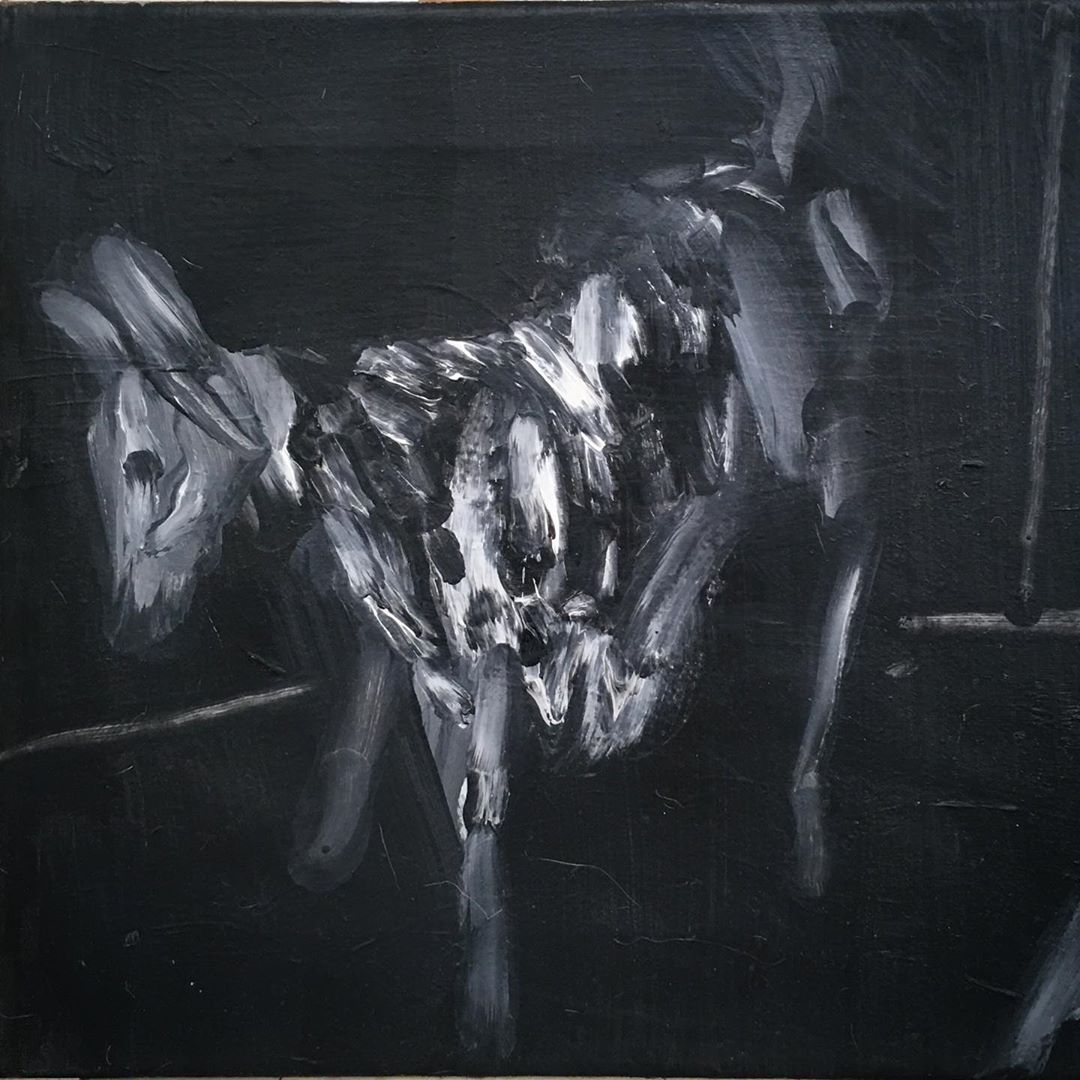}
        \caption{Albert Gouthier}
    \end{subfigure}\hfill
    \begin{subfigure}[t]{.24\textwidth}
        \centering\includegraphics[width=1.0\linewidth]{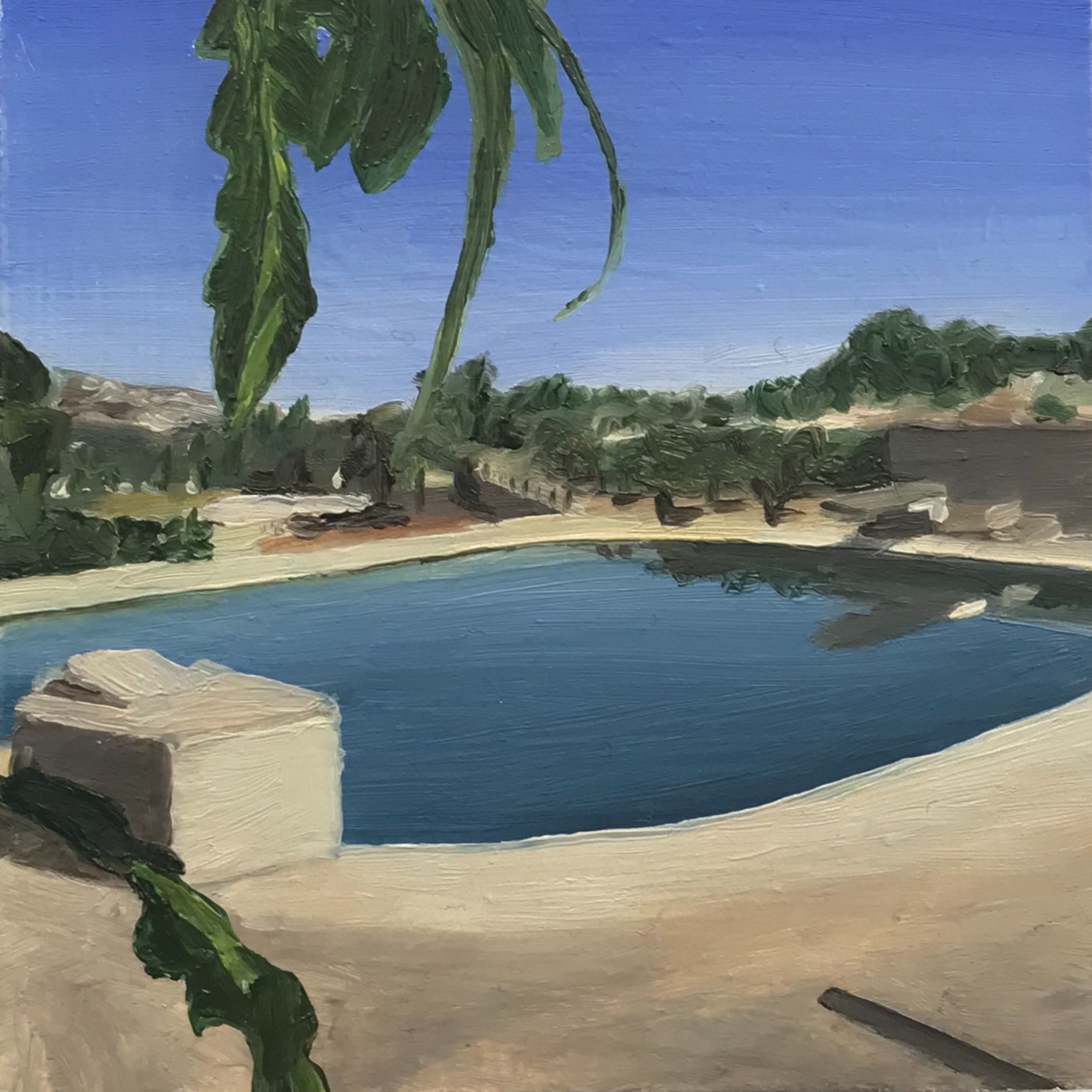}
        \caption{Janka Z{\"o}ller}
    \end{subfigure}\hfill
    \begin{subfigure}[t]{.24\textwidth}
        \centering\includegraphics[width=1.0\linewidth]{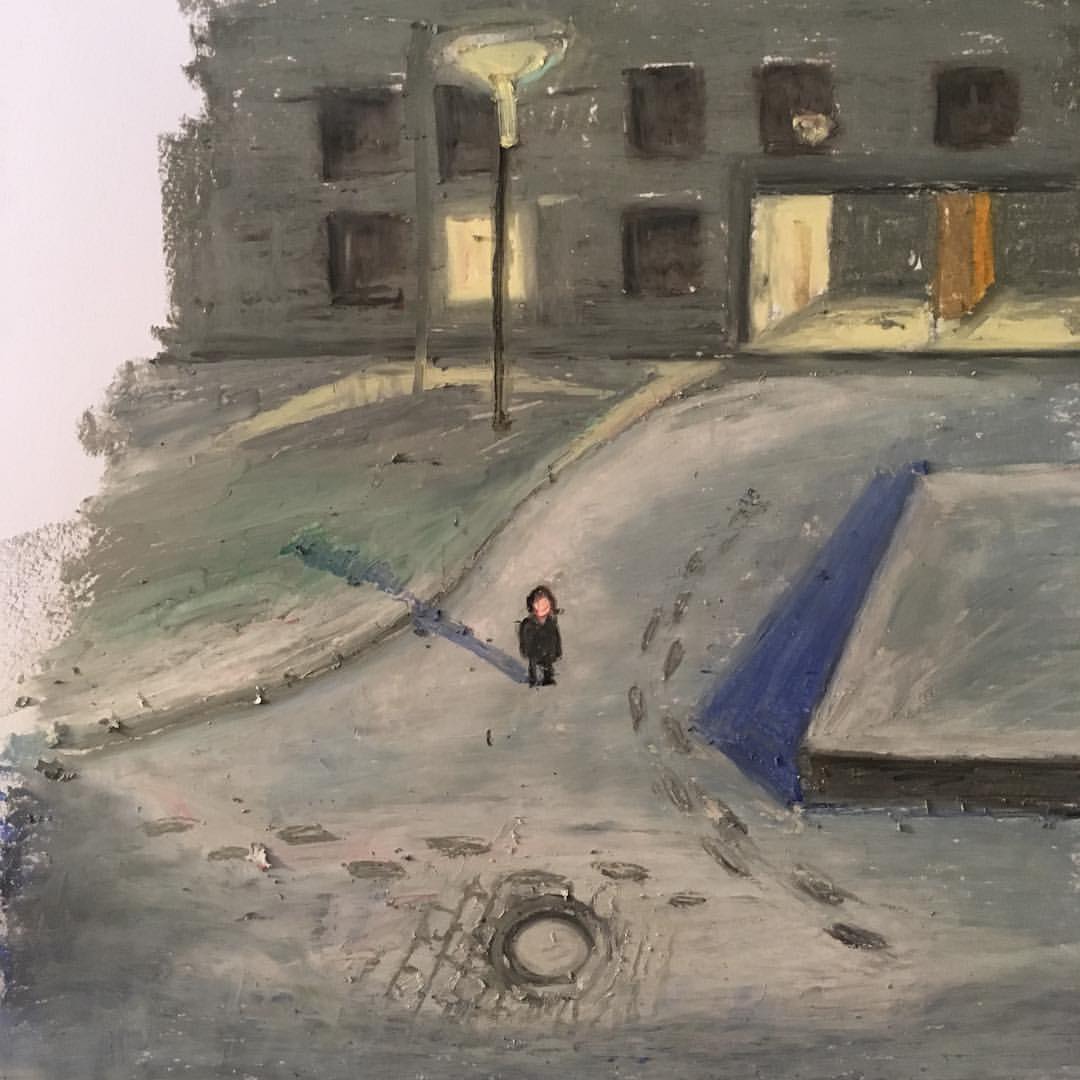}
        \caption{Soomin Kim}
    \end{subfigure}\hfill
    \begin{subfigure}[t]{.24\textwidth}
        \centering\includegraphics[width=1.0\linewidth]{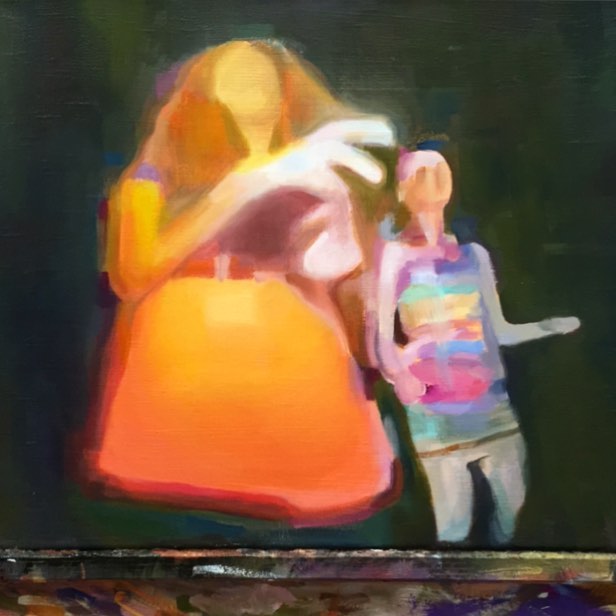}
        \caption{Allistair Walter}
    \end{subfigure}
    \\
    \begin{subfigure}[t]{.24\textwidth}
        \centering\includegraphics[width=1.0\linewidth]{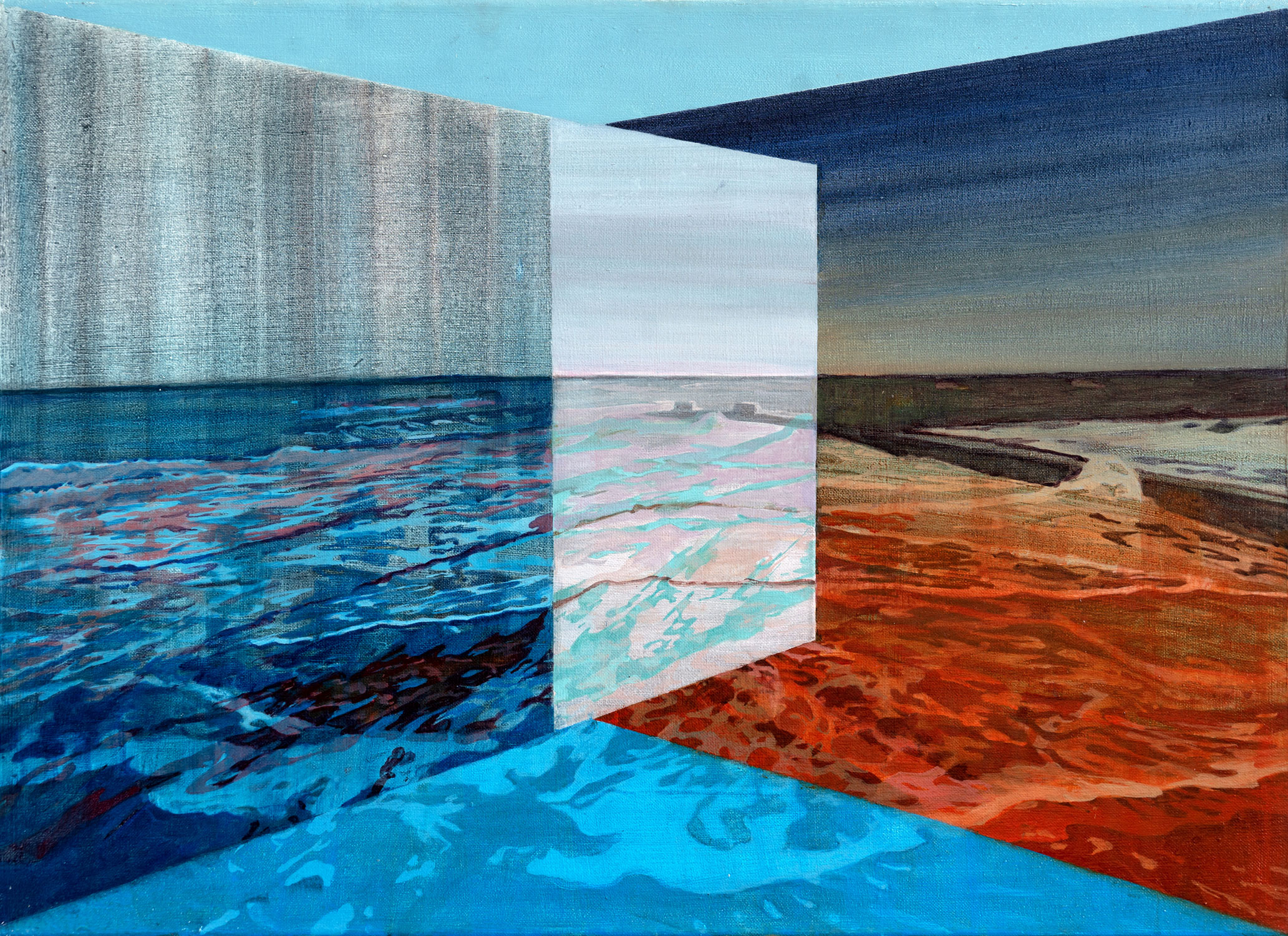}
        \caption{Soojie Kang}
    \end{subfigure}\hfill
    \begin{subfigure}[t]{.24\textwidth}
        \centering\includegraphics[width=1.0\linewidth]{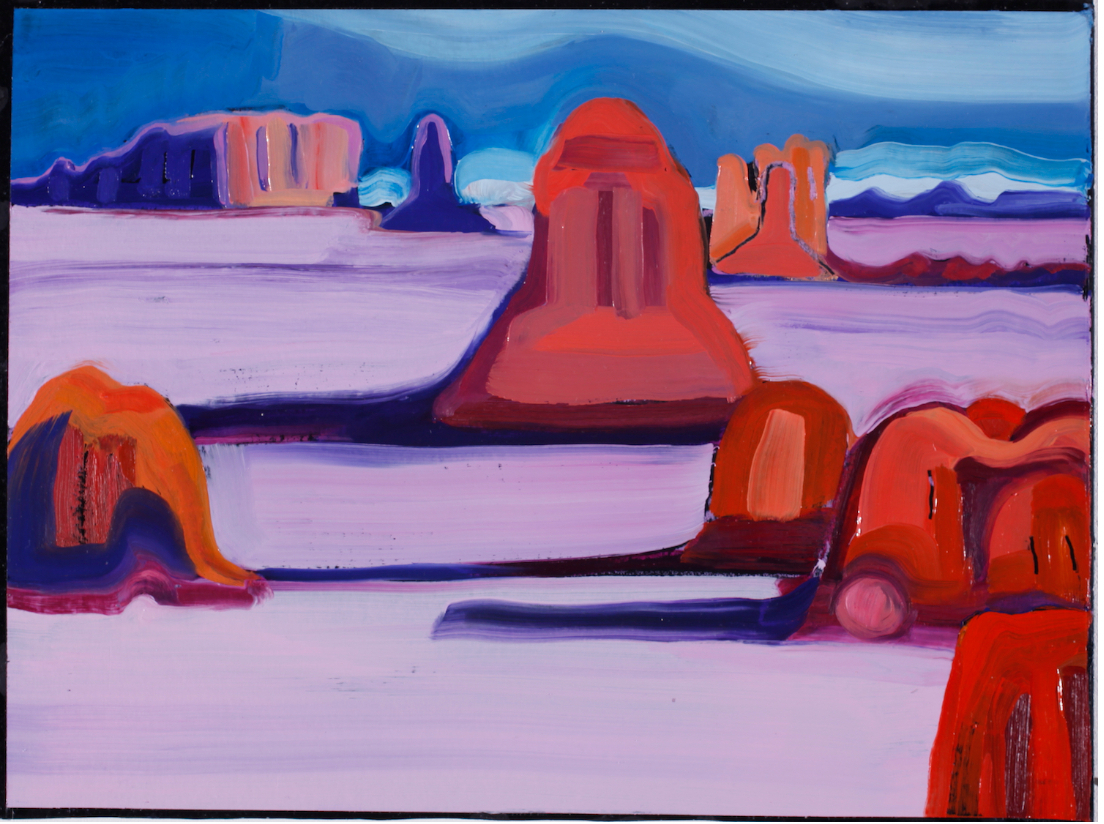}
        \caption{Olivia Parkes}
    \end{subfigure}\hfill
    \begin{subfigure}[t]{.24\textwidth}
        \centering\includegraphics[width=1.0\linewidth]{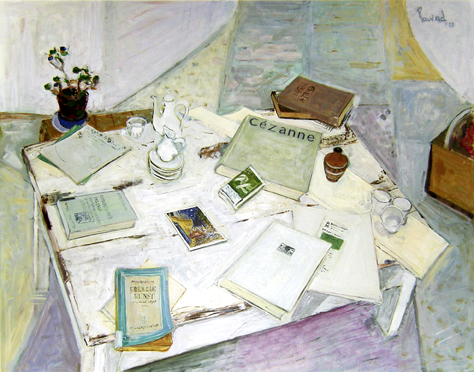}
        \caption{Rawad Atfeh}
    \end{subfigure}\hfill
    \begin{subfigure}[t]{.24\textwidth}
        \centering\includegraphics[width=1.0\linewidth]{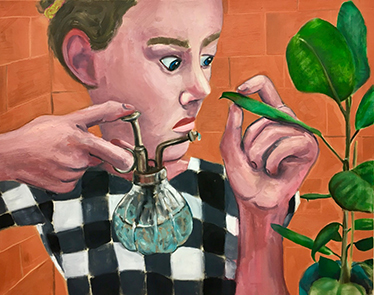}
        \caption{Marcel Kimble}
    \end{subfigure}
    \\
    \begin{subfigure}[t]{.24\textwidth}
        \centering\includegraphics[width=1.0\linewidth]{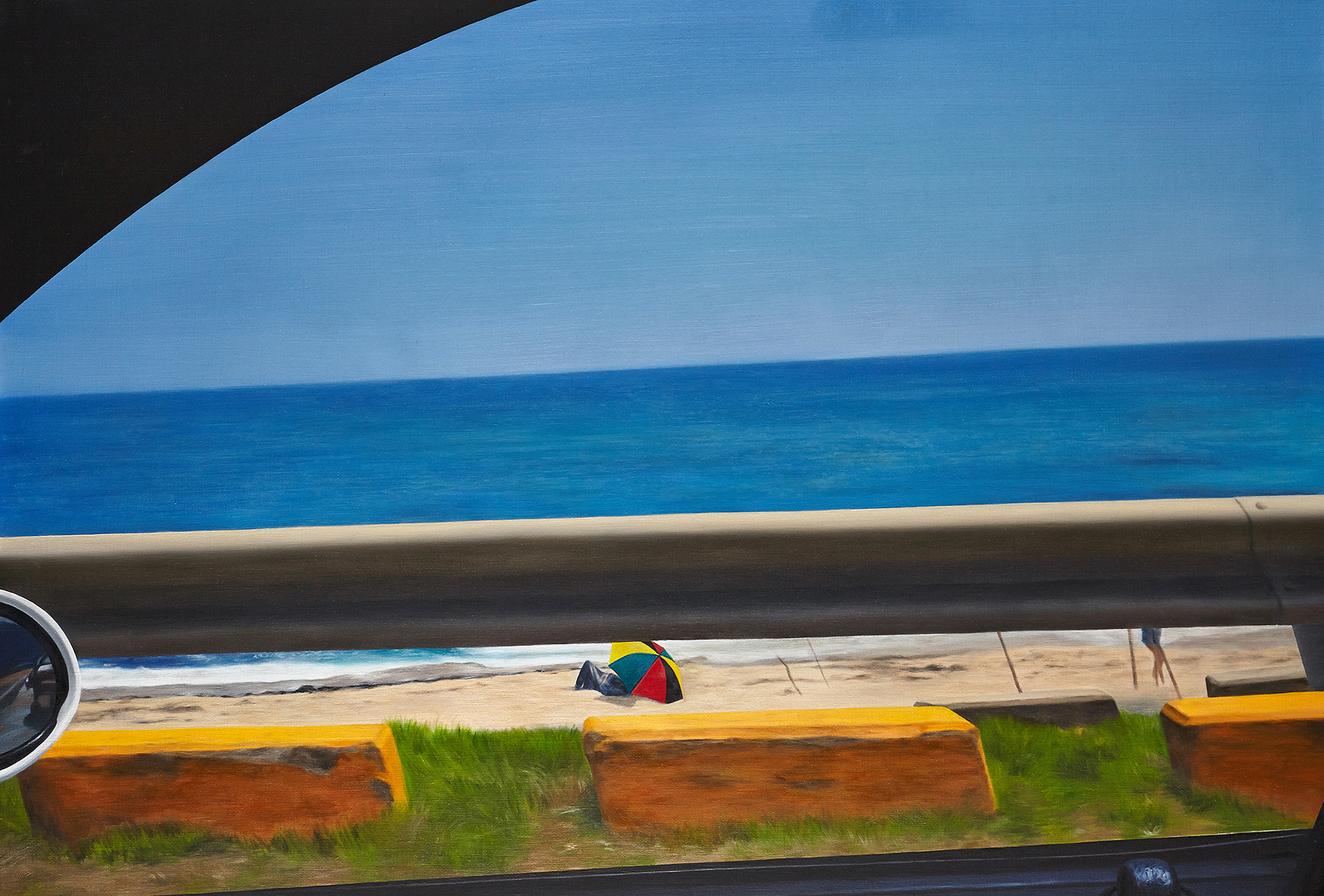}
        \caption{Jisub Kim}
    \end{subfigure}\hfill
    \begin{subfigure}[t]{.24\textwidth}
        \centering\includegraphics[width=1.0\linewidth]{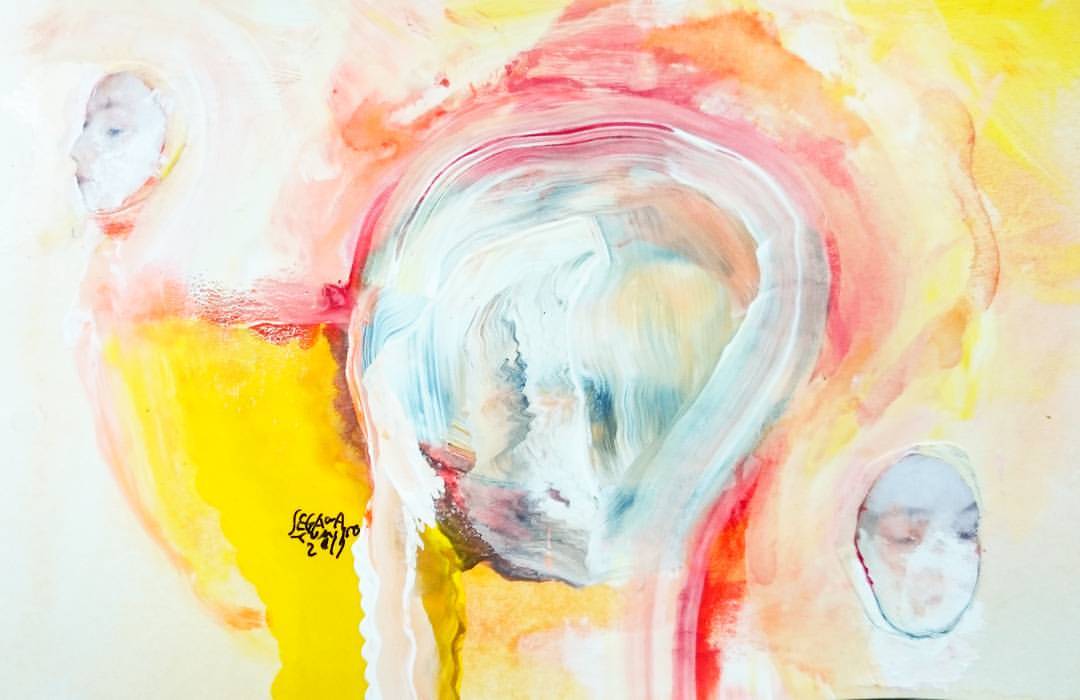}
        \caption{Yumiko Segawa}
    \end{subfigure}\hfill
    \begin{subfigure}[t]{.24\textwidth}
        \centering\includegraphics[width=1.0\linewidth]{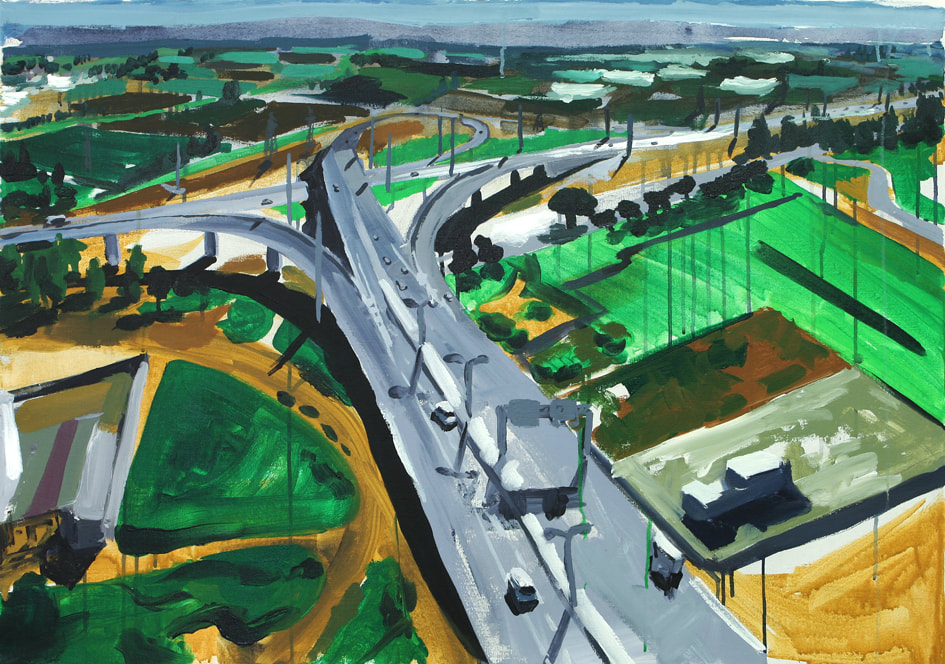}
        \caption{Ofra Ohana}
    \end{subfigure}\hfill
    \begin{subfigure}[t]{.24\textwidth}
        \centering\includegraphics[width=1.0\linewidth]{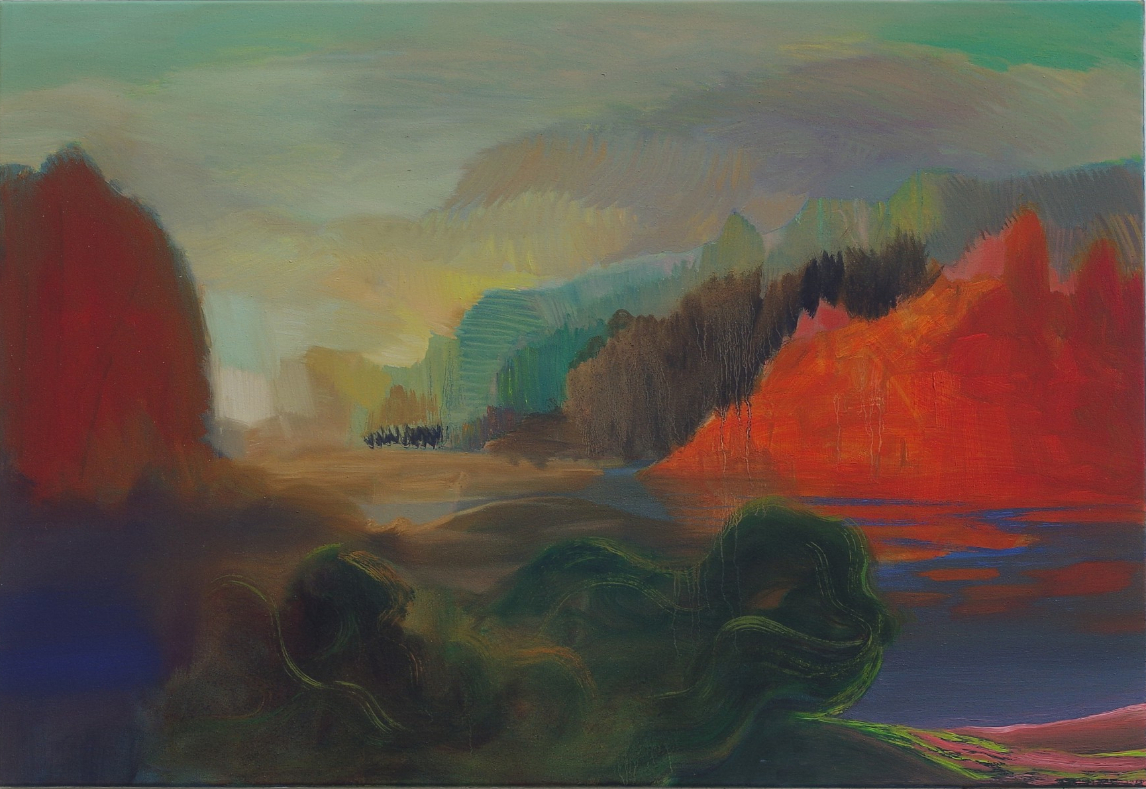}
        \caption{Martin Mischner}
    \end{subfigure}
    \caption{\textbf{contempArt image samples.}}\label{fig:examplePaint}
\end{figure}

\paragraph{Data collection.} Due to the manual and time-consuming nature of the data collecting process described in the following text, only art students in Germany were included in the analysis. To create the contempArt dataset, we first gather information on student enrolment in all fine arts programs related to painting or drawing at German art schools. This information is not publicly available until students join a specific painting or drawing class associated with one professor. These painting classes often have an online presence on which the names of current and former students are provided.\footnote{Example: \url{http://www.klasse-orosz.de/}} This data on student names, university and class membership was used to manually locate students individual websites as well as public Instagram\footnote{\url{https://www.instagram.com}} accounts, which were subsequently crawled in bulk for image content. If Instagram membership was known, further social media data was collected such as the students Instagram connections or detailed meta-data on the images posted on the network. Similarly, if artists webpages were found, self-reported biographical data on nationality and gender was collected. Art students who did not make images of their paintings available were omitted. Furthermore, any images not containing paintings or drawings were manually removed. Aggregate information on included art schools and their students can be seen in Table~\ref{tab:univ}; example images can be seen in Figure~\ref{fig:examplePaint}.

\begin{table}[t]
\setlength{\tabcolsep}{7pt}
\centering
\caption{\textbf{Data sources for contempArt.}}
\label{tab:univ}
\begin{tabular}{rlrr}
\toprule
ID & Art School & Students & Images \\ \midrule
1 & Alanus University of Arts and Social Sciences & 25 & 677 \\
2 & Wei{\ss}ensee Academy of Art Berlin & 8 & 144 \\
3 & Berlin University of the Arts & 24 & 601 \\
4 & Braunschweig University of Art & 39 & 1,122 \\
5 & University of the Arts Bremen & 29 & 991 \\
6 & Dresden Academy of Fine Arts & 44 & 1,743 \\
7 & Burg Giebichenstein University of Art and Design & 18 & 777 \\ 
8 & Hochschule f{\"u}r Grafik und Buchkunst Leipzig & 68 & 2,623 \\ 
9 & Mainz Academy of Arts & 19 & 427 \\
10 & Academy of Fine Arts M{\"u}nchen & 44 & 1,227 \\ 
11 & Kunstakademie M{\"u}nster & 33 & 2,238 \\
12 & Academy of Fine Arts N{\"u}rnberg & 37 & 555 \\
13 & Hochschule f{\"u}r Gestaltung Offenbach am Main & 11 & 191 \\
14 & Hochschule der Bildenden K{\"u}nste Saar & 25 & 553 \\
15 & State Academy of Fine Arts Stuttgart & 18 & 690 \\ \midrule
\end{tabular}%
\end{table}

\begin{figure}[t]
    \begin{subfigure}[t]{0.56\textwidth}
        \includegraphics[width=\textwidth]{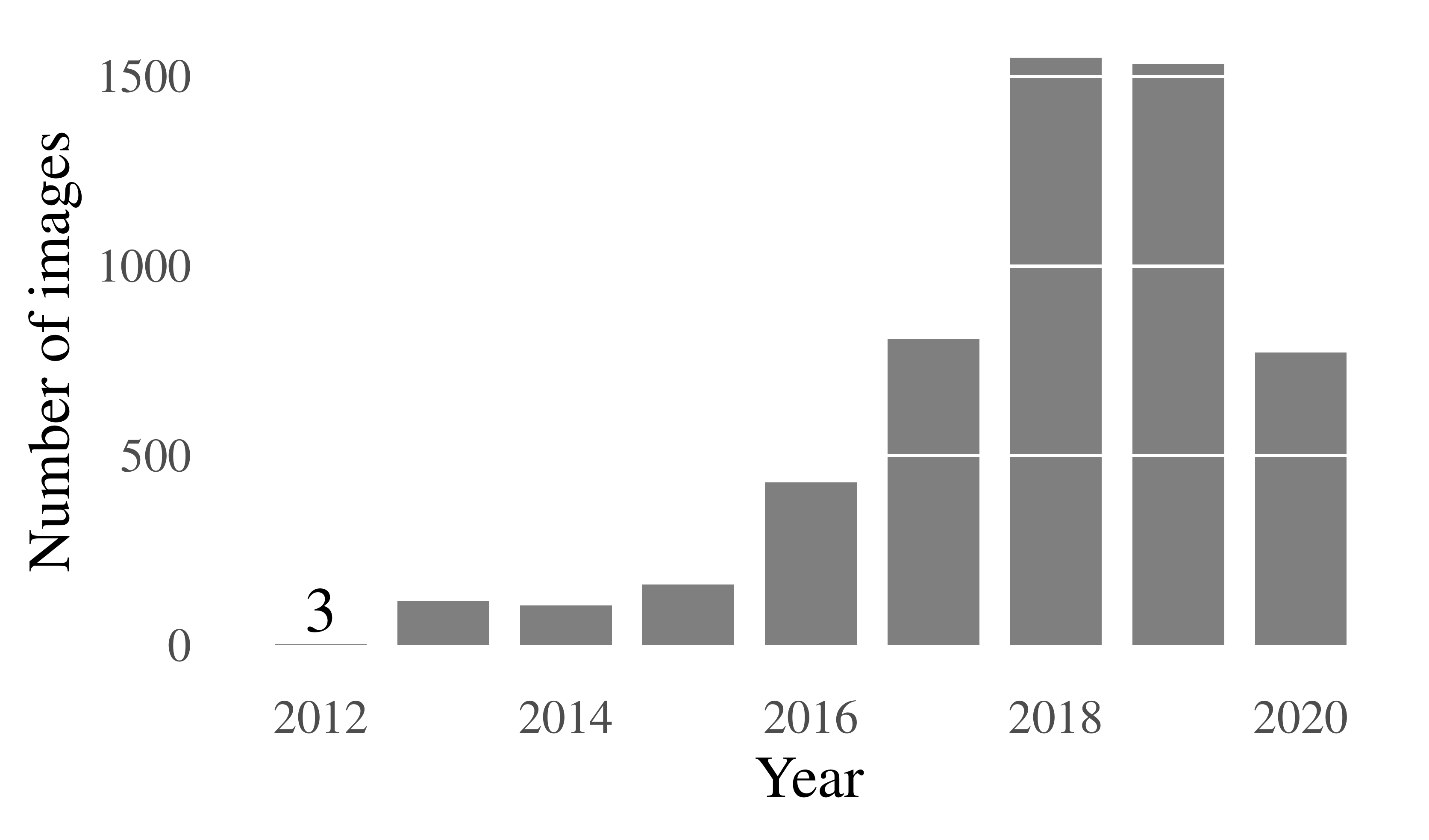}
    \end{subfigure}\hfill
    \begin{subfigure}[t]{0.4\textwidth}
        \includegraphics[trim={0.20cm 0.20cm 0.15cm 0.15cm},clip,width=\textwidth]{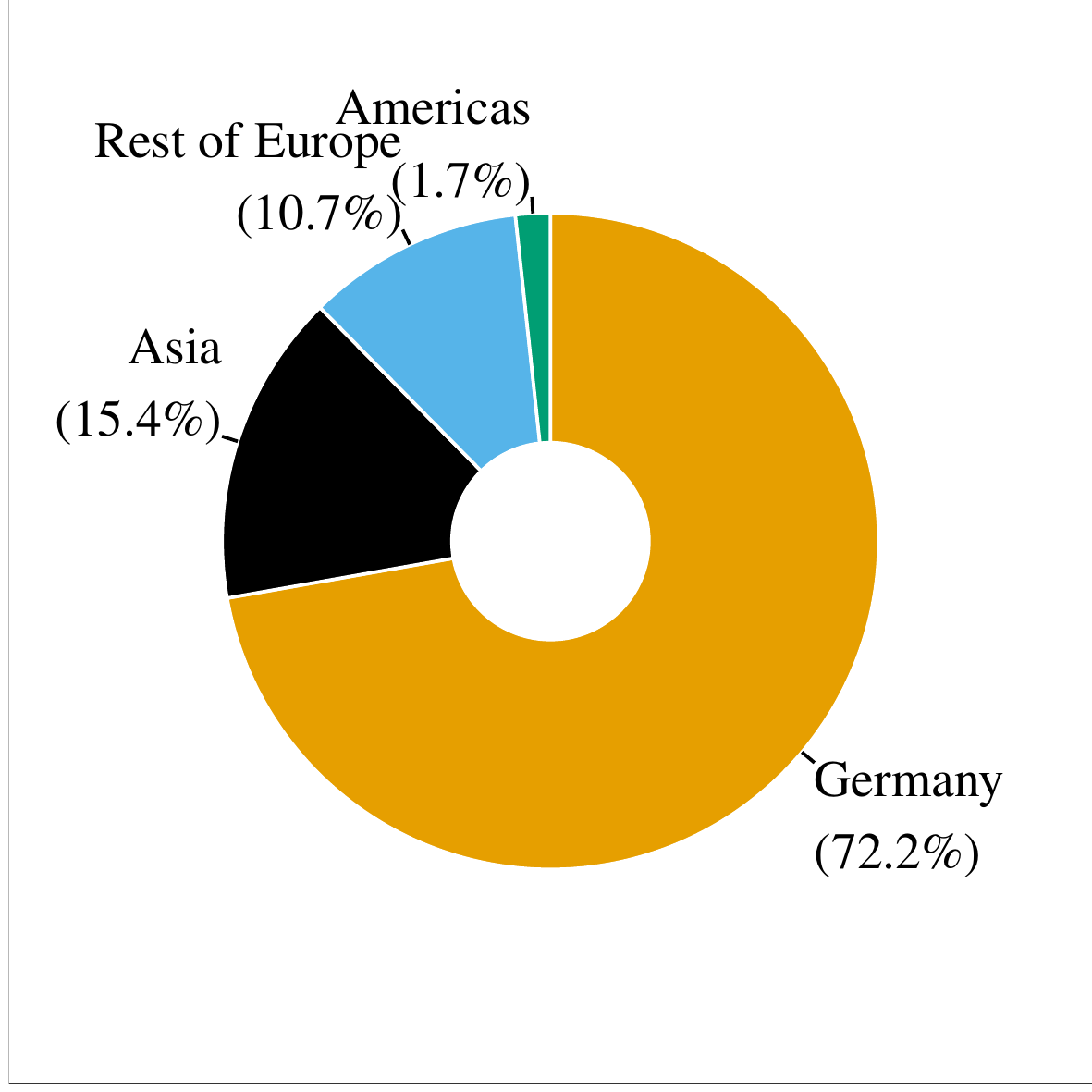}
    \end{subfigure}
    \caption{\textbf{Metadata distribution for contempArt}. Timeframe only available for 5,478 Instagram images. Nationality only available for 234 artists.}
    \label{fig:datDescr}
\end{figure}

\paragraph{Dataset statistics.} Less than half of the original list of 1,177 enrolled students had any findable online presence. From the final set of 442 artists, 14,559 images could be collected, with the median number per artists being 20. The data sources were both Instagram accounts and webpages, whereas 37.78\% of students only had the former, 17.19\% only had the latter, and 45.02\% had both. Each data source contributed different metadata to the dataset. Dedicated homepages, the source for 62.37\% of all images, generally contain self reported information on the artists nationality and gender. The image data from Instagram, on the other hand, was of lower quality\footnote{Images available on artists dedicated webpages are generally of high resolution and only depict their work. Contrary to Instagram, which limits the image resolution by default to $1080\times1080$ pixels and where the images uploaded by the artists were often noisy; e.g. taken from a larger distance or of artwork surrounded by objects. Cropping away unnecessary content further reduced the image size.} but contained time annotations that allow the estimation of a time range in which the collected artworks were produced. The distribution of images over time and nationality of artists is shown in Figure~\ref{fig:datDescr}. Most importantly, the Instagram account metadata provides detailed information on the social media connections between the artists.

\begin{figure}[t]
    \centering
    \includegraphics[width=\textwidth]{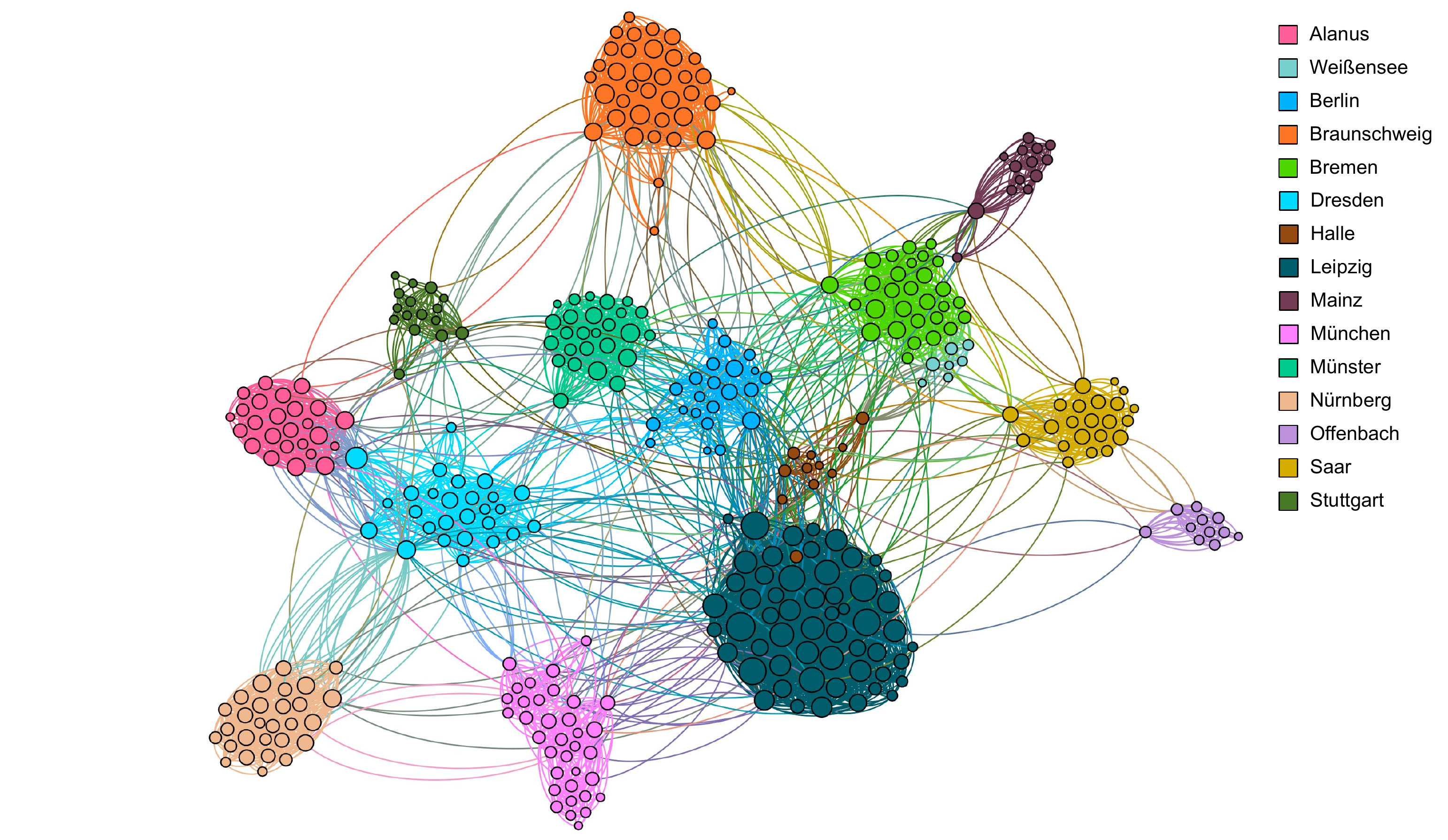}
    \caption{\textbf{Visualisation of} $\mathfrak{S}^{\text{U}}$ \textbf{with Gephi \cite{gephi}}. Each node is an artist and the colouring is mapped to art school affiliation. The direction of edges is clockwise and node size represents the number of edges per node.
    }\label{fig:artistNetwork}
\end{figure}

\paragraph{Instagram network graphs.} The sets of Instagram accounts following and being followed by each artist, available for 82.35\% of the sample,\footnote{Two Instagram accounts were deleted or renamed during the data collection process so only their image data is available.} allow the construction of two directed network graphs capturing information on social relationships. We denote the directed artist network as $\mathfrak{S}^{\text{U}}=(O^{\text{U}},D^{\text{U}})$, where $O^{\text{U}}$ is the set of artists, node $o^{\text{U}}_i \in O^{\text{U}}$ denotes artist $i$, and edge $d^{\text{U}}_{ij} \in D^{\text{U}}$ denotes artist $i$ following artist $j$. In $\mathfrak{S}^{\text{U}}$, the number of nodes is 364, the number of edges 5,614 and the median number of edges per node 27. This network, visualised in Figure~\ref{fig:artistNetwork}, is closely related to art school membership - only 9.73\% of edges are between artists of different schools - and to a smaller extent geography, e.g. the proximity of Halle, Leipzig and Berlin. Art schools serve as the primary social hub for most young students and nationwide connections only become common in later stages of their career, so this is to be expected. We denote a second, unconstrained directed network by $\mathfrak{S}^{\text{Y}}=(O^{\text{Y}},D^{\text{Y}})$, where $O^{\text{Y}}$ is the set of all Instagram accounts followed by or following artists. The node $o^{\text{Y}}_k \in O^{\text{Y}}$ denotes account $k$, and edge $d^{\text{Y}}_{kl} \in D^{\text{Y}}$ denotes account $k$ following account $l$. The number of nodes in $\mathfrak{S}^{\text{Y}}$ is 247,087 and the number of edges is 745,144.

\section{Unsupervised Style Embeddings}

In order to compute image embeddings that are closely related to artistic style, we follow three established, unsupervised approaches that are all based on the VGG network for image classification~\cite{simonyan2014very}. Although newer and deeper CNN, such as ResNet~\cite{he2016deep}, have since been proposed that outperform the VGG-network on its original task, it has become a widely used tool in both art classification tasks~\cite{mao2017deepart} and texture synthesis~\cite{gatys2016image,jing2019neural}. After presenting the different methods, we will examine their visual and numerical connection to labels and images of a commonly used fine art dataset.

\paragraph{Raw VGG embeddings.} We use the deepest network variant of VGG with 19 stacked convolutional layers and three fully connected layers on top. The network is pre-trained on the ImageNet database~\cite{deng2009imagenet} and the second to last layer $\text{fc}_7$ is used as the style embedding $\mathbf{e}_n^\mathrm{V}\in\R^{4,096}$ for any image $I$. Similar deep features~\cite{karayev2013recognizing}, that are derived from CNNs trained on ImageNet and not art in particular, have been shown to contain salient information about the latter.

\paragraph{Texture-based VGG embeddings.} In the seminal work of~\cite{gatys2016image} it has been shown that deep CNNs, and VGG in particular, can be leveraged to perform arbitrary artistic style transfer between images. Specifically, that the correlations inside convolutional feature maps of certain network layers capture positionless information on the texture or rather, the style of images. This, so-called, Gram-based style representation has been widely used in texture synthesis~\cite{jing2019neural} and art classification~\cite{mao2017deepart,chu2018image}. Contrary to~\cite{gatys2016image}, in which this style representation is a part of an optimisation procedure aligning the texture of two images, we utilise it only as a further embedding of style $\mathbf{e}_n^\mathrm{T}$. The extraction process is as follows:

Consider the activations at feature map $\textbf{F}_{\ell}(I)\in \R^{C_{\ell}\times(H_{\ell}W_{\ell})}$ of image $I$ at layer $\ell$ of the VGG network described in the previous subsection. $C_{\ell}$ is the number of channels, and $W_{\ell}$ and $H_{\ell}$ represent the width and height of feature map $\textbf{F}_{\ell}(I)$. $\textbf{F}_{\ell}(I)[j]$ denotes the column vector in $\R^{C_{\ell}}$ that holds the feature map activations at pixel position $j \in \{1,\dots,H_{\ell}W_{\ell}\}$. Following the proposed normalisation procedure in~\cite{li2017diversified}, the Gram matrix $\textbf{G}_{\ell} \in\R^{C_{\ell}\times C_{\ell}}$ of the \textit{centered} feature maps at $\ell \in L = \{conv_{1\_1},conv_{2\_1},conv_{3\_1},conv_{4\_1},conv_{5\_1}\}$, given by
\begin{equation}
    \textbf{G}_{\ell}=\frac{1}{H_{\ell}W_{\ell}}\sum_{j=1}^{H_{\ell}W_{\ell}}(\textbf{F}_{\ell}(I)[j]-\pmb{\mu}_{\ell})(\textbf{F}_{\ell}(I)[j]-\pmb{\mu}_{\ell})^\top,
\end{equation}
 and the means $\pmb{\mu}_{\ell} \in\R^{C_{\ell}}$ themselves, given by
\begin{equation}
    \pmb{\mu}_{\ell}=\frac{1}{H_{\ell}W_{\ell}}\sum_{j=1}^{H_{\ell}W_{\ell}} \textbf{F}_{\ell}(I)[j],
\end{equation}
are concatenated into a high-dimensional collection $\{\pmb{\mu}'_\ell,\textbf{G}'_\ell|\ell \in L\}$ with further normalisation by
\begin{equation}
    \pmb{\mu}'_{\ell} =  \frac{\pmb{\mu}_{\ell}} {C_{\ell}(C_{\ell}+1)} \;\;\;\;\; \textbf{G}'_{\ell} =  \frac{\textbf{G}_{\ell}} {C_{\ell}(C_{\ell}+1)},
\end{equation} 
in line with~\cite{wynen2018unsupervised}. The Gram matrix $\textbf{G}_{\ell}$ is symmetric, so values below the diagonal are omitted. 
The collection is vectorised to a $S = \sum_\ell C_\ell (C_\ell + 3)/2$-dimensional texture descriptor $\mathbf{v}$, which can be computed for any image $I$. However, due to $\textbf{v}$ being very high-dimensional it is common practice to apply a secondary dimensional reduction on the joint $N$ texture descriptors of the present image dataset~\cite{gatys2015texture,wynen2018unsupervised,matsuo2016cnn}. To do so, we aggregate $\textbf{v}_n$ for all images $n=1,\ldots,N$ in the given dataset and concatenate them into a matrix $\mathbf{V} = [\mathbf{v}_i, \dots, \mathbf{v}_N] \in \R^{N\times S}$. We apply singular value decomposition to this matrix, extracting 4,096-dimensional features as our second style embedding $\mathbf{e}_n^\mathrm{T}\in\R^{4,096}$ for image $I_n$.

\paragraph{Archetype embeddings.} Wynen et al.~\cite{wynen2018unsupervised} uses the previously described Gramian texture descriptor $\mathbf{e}^\mathrm{T}$ and a classical unsupervised learning method called archetypal analysis \cite{cutler1994archetypal} to compute and visualise a set of art archetypes.\footnote{Definition of archetype: the original pattern or model of which all things of the same type are representations or copies~\cite{MerriamWebster2009}.} With archetypal analysis, the $P$ $K$-dimensional samples of an original matrix $\textbf{X}=[\textbf{x}_1,\dots,\textbf{x}_P]$ are approximately reconstructed as convex mixtures of $M$ archetypes\linebreak[4]$\mathbf{Z}^\top = [\mathbf{z}_1,\dots,\mathbf{z}_M] \in \mathbb{R}^{K \times M}$, i.e.,
\begin{equation}
    \textbf{x}_p \approx \textbf{Z} \pmb{\alpha}_p,\quad\text{with}\quad\sum_{m=1}^M\alpha_{pm}=1,\quad \alpha_{pm} \geq 0
\end{equation}
where $\pmb{\alpha}_n\in\R^M$ contain the mixture coefficients, $\alpha_{pm}$'s, that approximate each $p=1,\dots,P$ observations by a combination of archetypes, whereas the \linebreak[4]$m=1,\dots,M$ archetypes are themselves convex mixtures of samples:
\begin{equation}
    \textbf{z}_m=\textbf{X}\pmb{\beta}_m,\quad\text{with}\quad\sum_{p=1}^P\beta_{mp}=1,\quad \beta_{mp} \geq 0
\end{equation}
where $\pmb{\beta}_m \in \R^P$ contain the mixture coefficients, $\beta_{mp}$'s, that approximate each archetype with $\mathbf{X}$. For ease of notation, let $\mathbf{A}\in\R^{M\times P}$ and $\mathbf{B}\in\R^{P\times M}$ be matrices that contain $\pmb{\alpha}_p$'s and $\pmb{\beta}_m$'s, respectively. Then, the optimal weights of $\textbf{A}$ and $\textbf{B}$ can be found by minimising the residual sum of squares 
\begin{equation}
    \text{RSS}_M =  \parallel\textbf{X}-\textbf{XBA}\parallel_F^2,
\end{equation}
subject to the above constraints, with efficient solvers \cite{chen2014fast}. The number $M$ of archetypes can be predefined, as in~\cite{wynen2018unsupervised}, or adjusted by visually comparing $\text{RSS}_M$ at different $M$-values as in the original work~\cite{cutler1994archetypal}. We apply archetypal analysis on the matrix of stacked texture descriptors $\textbf{E}^\mathrm{T}=[\mathbf{e}_1^\mathrm{T},\dots,\mathbf{e}_N^\mathrm{T}]\in\R^{N\times4,096}$ for the given dataset containing $N$ images. The estimated archetype-to-image and image-to-archetype mixture weights $\pmb{\alpha}_n$ and $\pmb{\beta}_n^T$ of each image $I_n$ are then concatenated into the final style embedding $\mathbf{e}_n^\mathrm{A}\in\R^{2M}$.

\section{Comparative Evaluation of Style Embeddings}

The unsupervised nature of the described embeddings - subsequently called \textit{VGG}, \textit{Texture} and \textit{Archetype} - requires an evaluation of their connection to artistic style and differences therein. Due to the visual nature of artworks, evaluations of their unsupervised feature spaces often rely only on visual comparisons, as in the case of the archetypal style embeddings~\cite{wynen2018unsupervised} or texture synthesis~\cite{yeh2020improving,jing2019neural,kotovenko2019content}. In the following, we investigate both the visual differences between the embeddings as well as their relation to existing style labels. 

\paragraph{Evaluation details.} We download a balanced subset of Wikiart, sampling 1,000 random images each from the 20 most frequent style labels after excluding non-painting artworks that were classified as photography, architecture, etc.\footnote{Included styles: Abstract Art, Abstract Expressionism, Art Informel, Art Nouveau (Modern), Baroque, Cubism, Early Renaissance, Expressionism, High Renaissance, Impressionism, Naïve Art (Primitivism), Neoclassicism, Northern Renaissance, Post-Impressionism, Realism, Rococo, Romanticism, Surrealism, Symbolism, Ukiyo-e.} The VGG network is initialised with its pre-trained weights for image classification and this same network is used to generate all three embeddings. Images are scaled down to 512 pixels on their shorter side while maintaining aspect ratio and then center-cropped to $512\times512$ patches. For the \textit{Archetype} embedding, the number of archetypes was set to $M=40$ - twice the number of style labels~- although further empirical comparisons showed that varying $M$ had only marginal influence on either of the results. 
\begin{figure}[t]
    \centering
    \includegraphics[width=\textwidth]{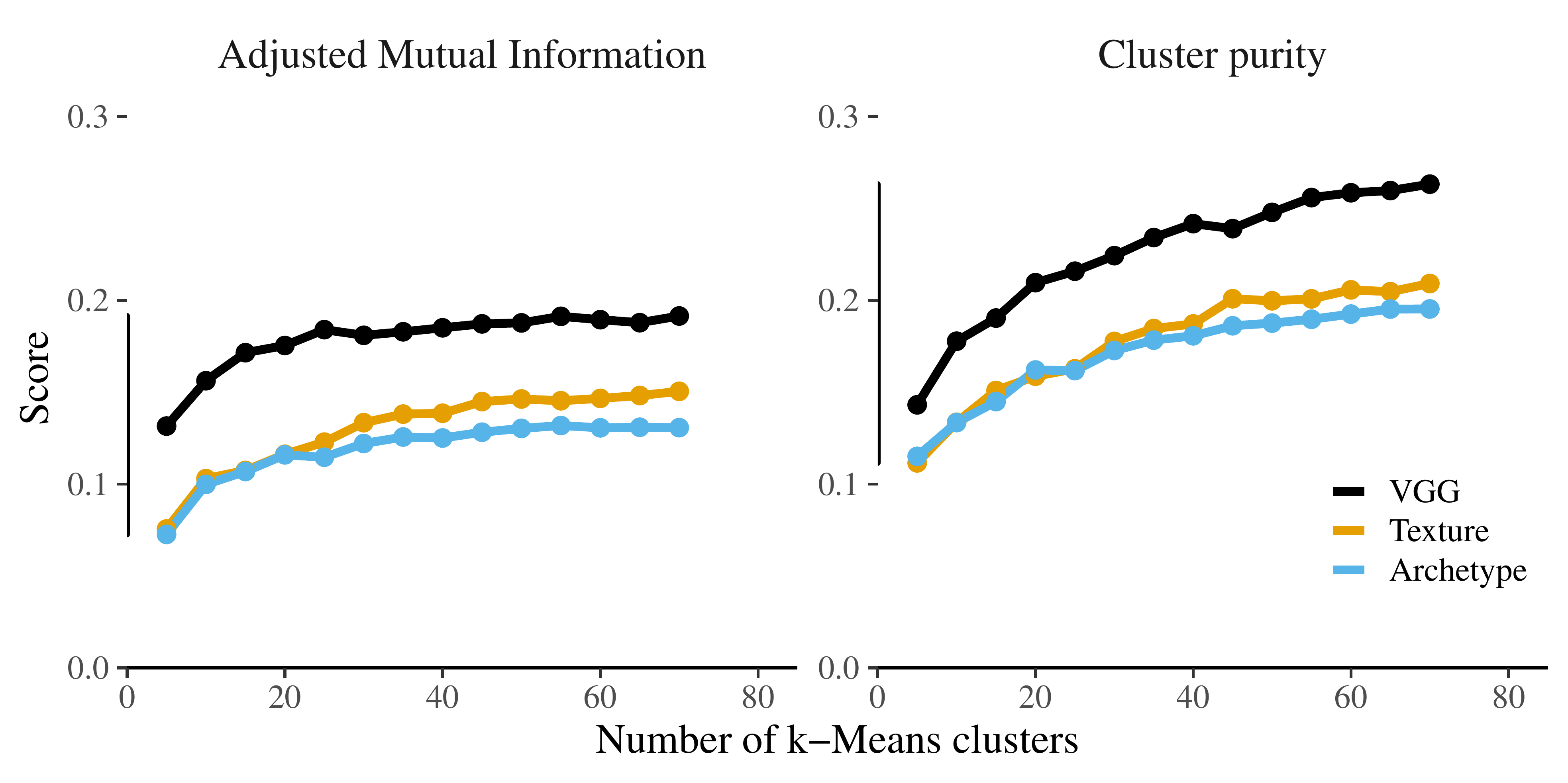}
    \caption{\textbf{Evaluation of style embeddings.} Similarity between k-Means clustering based on the three unsupervised embeddings and style labels from Wikiart.}\label{fig:evalNum}
\end{figure}
\paragraph{Numerical evaluation.} For a range of $k=5,10,15,\dots,70$, we partition the three embeddings into $k$ clusters with the k-Means algorithm. The informational overlap between the resulting cluster memberships and the existing style annotations is calculated with the adjusted mutual information (AMI) score~\cite{vinh2010information}, a normalised and chance-adjusted metric that quantifies the reduction in class entropy when the cluster labels are known. A value of 1 would represent cluster and class membership matching up perfectly, whereas values around 0 would signify a random clustering. In order to provide a more transparent yet unnormalised evaluation measure, we additionally show the purity score, for which the images in each cluster are assigned to its most frequent style label and the average number of thereby correctly assigned images is calculated. The results in Figure~\ref{fig:evalNum} show that the \textit{VGG} embeddings have the highest AMI and purity score for all values of $k$. The \textit{Archetype} and \textit{Texture} embeddings have similar results, even though the dimensionality of the former is 50 times less. Even the highest AMI-score of $0.191$ can still be considered closer to a random clustering than an informative one, leading to the conclusion that none of the embeddings correspond closely to commonly used labels of artistic style. However, style annotations in fine art datasets are known to be very broad and, in Wikiart's case, noisy~\cite{elgammal2018shape}, allowing for some margin of error and calling for a further, visual inspection of the embeddings.

\begin{figure}[H]
    \centering
    \includegraphics[width=0.8\textwidth,valign=c]{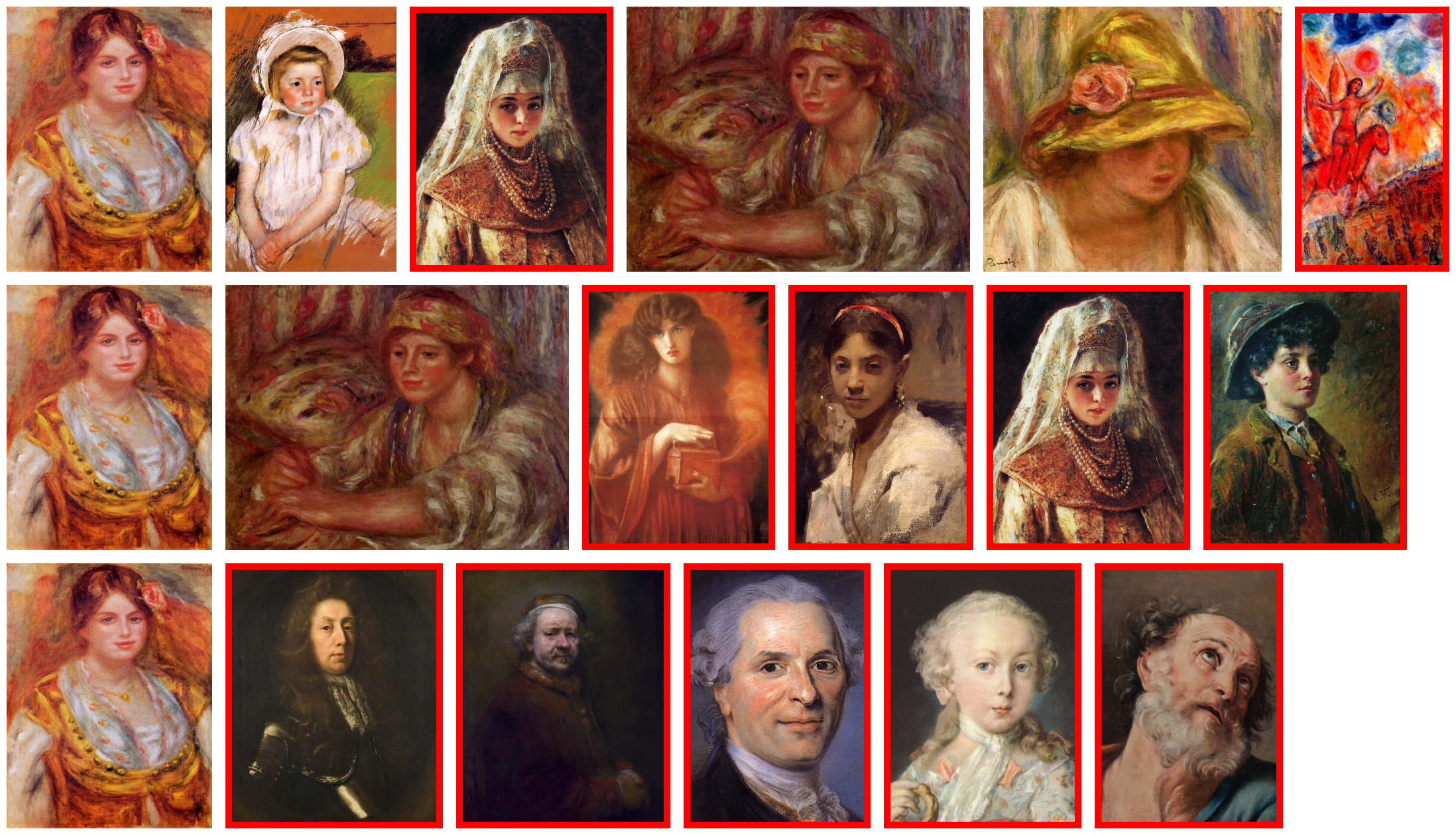}
    \\[\smallskipamount]
    \includegraphics[width=0.8\textwidth,valign=c]{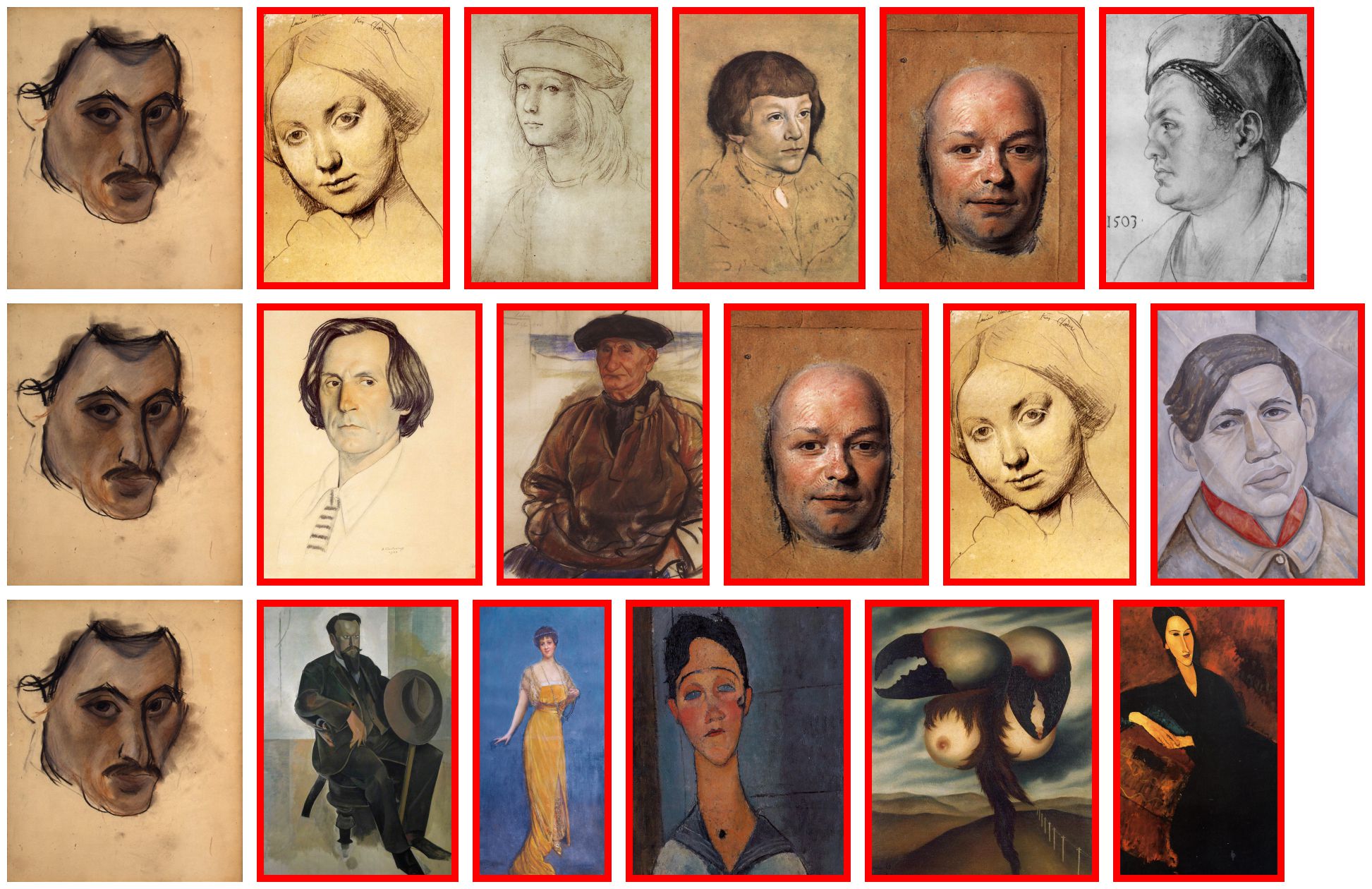}
    \\[\smallskipamount]
    \includegraphics[width=0.8\textwidth,valign=c]{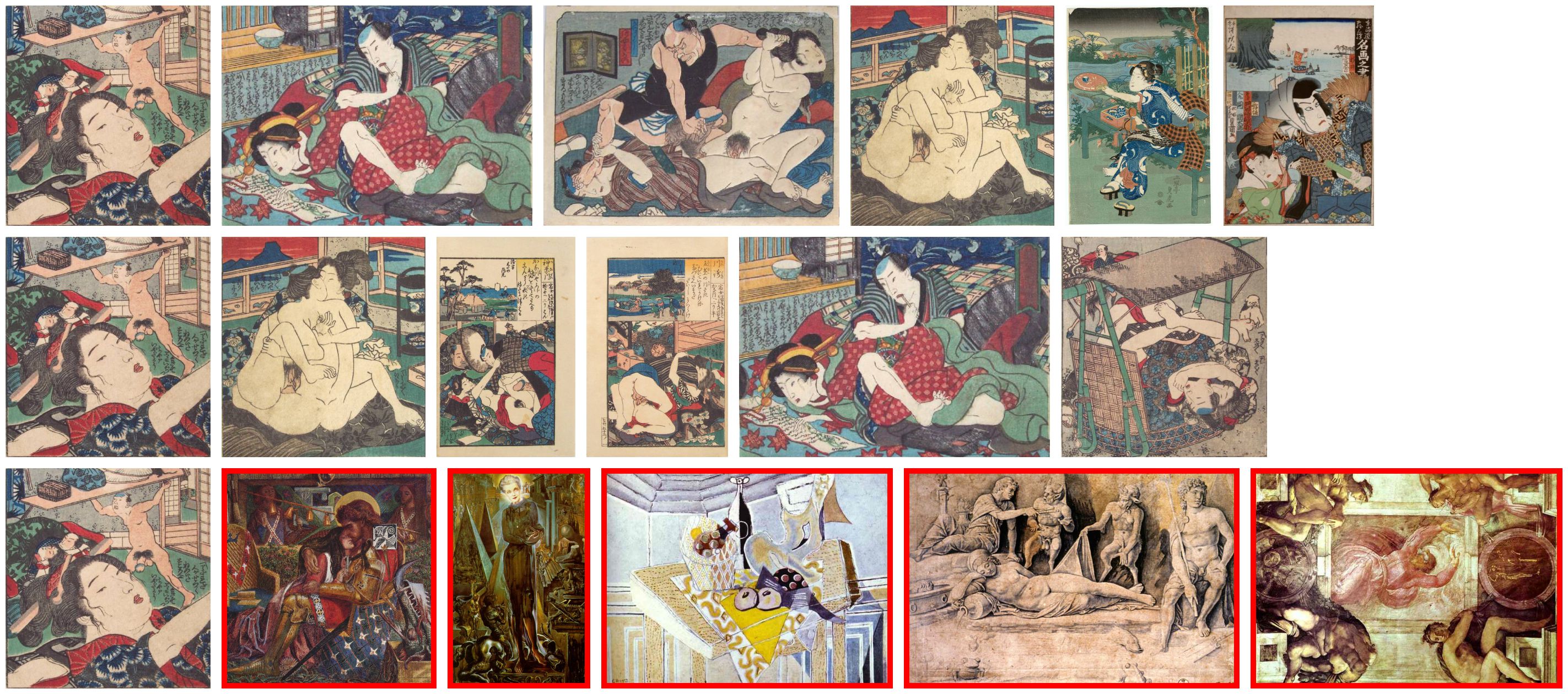}
    \caption{\textbf{Visual evaluation of style embeddings.} For three images, repeated throughout the first column, the five most similar images in all three style embeddings are shown in descending order from left to right. From top to bottom: \textit{VGG}, \textit{Texture} and \textit{Archetype}. A red border indicates that the chosen image does not share the anchor images style annotation from Wikiart.}\label{fig:evalVisual}
\end{figure}

\paragraph{Visual evaluation.} We visualise a small set of randomly chosen images with their five closest neighbours for each of the style embeddings. Closeness is calculated with the cosine similarity. The comparison in Figure~\ref{fig:evalVisual} gives insights into the difference between style annotations and stylistic similarity. The \textit{Archetype} embedding, not being able to cluster the visually unique Ukiyo-e genre as well as failing to align even general colour space, again performs the worst. Archetypal analysis, while allowing a high degree of interpretability and aesthetic visualisations \cite{wynen2018unsupervised,chen2014fast} by encoding images as convex mixtures of themselves, has to be evaluated more rigorously to validate its usefulness for art analysis. \textit{VGG} and \textit{Texture} are each able to match up images in terms of a general visual style. However, both are inconsistent in doing the same for labelled style, echoing the results of the numerical evaluation.

The overlap between the evaluated embeddings and regular style annotations was shown to be minimal, but two of the three still contain valid information on artistic visual similarity. Texture, although exceptional in transferring itself, does not capture style in the art historical sense. Conversely, that same style can not be described by visual content alone, validating context-based approaches to art classification tasks as in~\cite{garcia2019context}.

\section{Analysis of contempArt}
\begin{figure}[t]
    \centering
    \includegraphics[width=\textwidth]{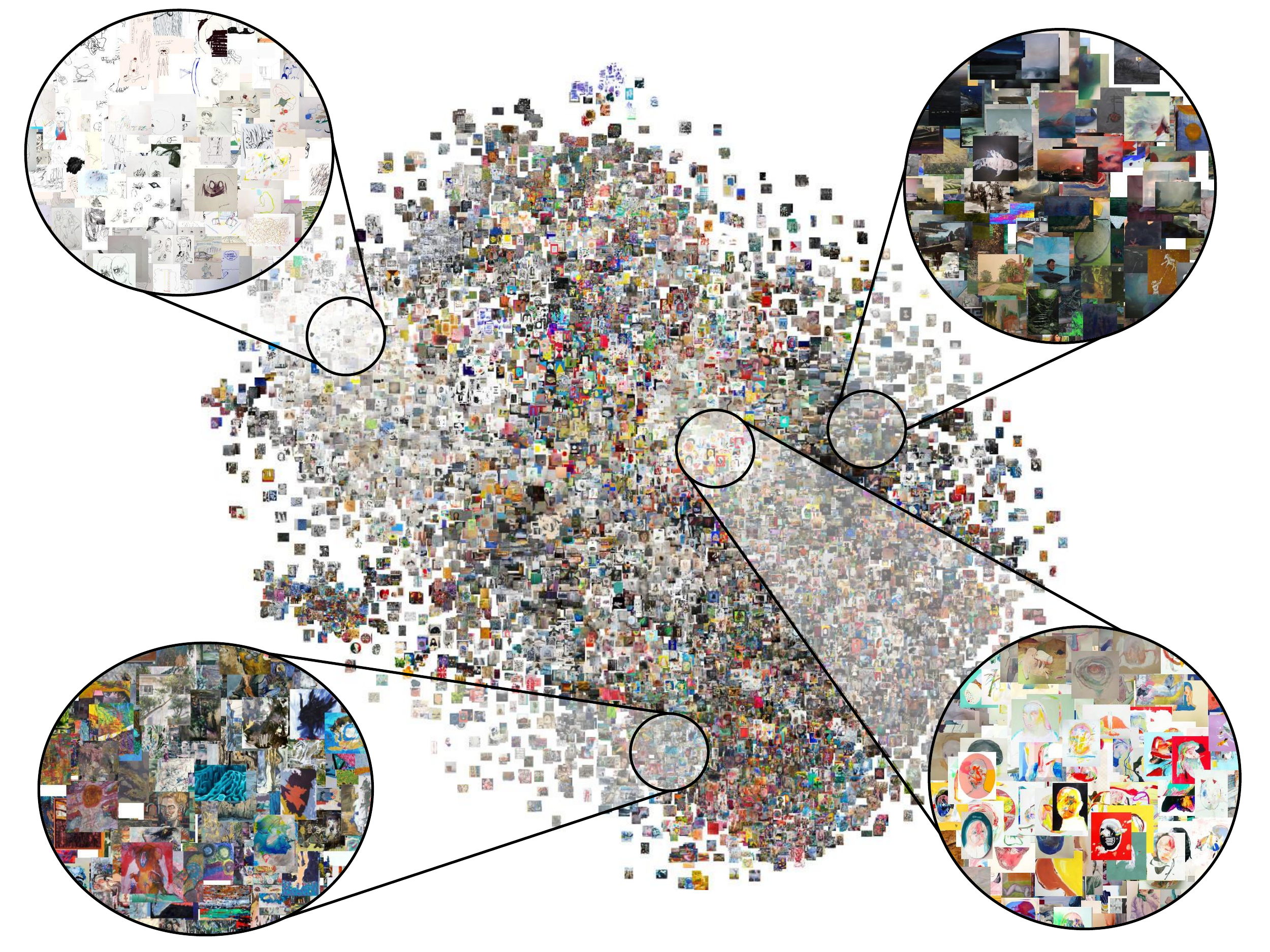}
    \caption{\textbf{Visualisation of contempArt with t-SNE~\cite{maaten2008visualizing}.}}
    \label{fig:tsnePainting}
\end{figure}

The VGG embeddings of the contempArt images, partially visualised in Figure~\ref{fig:tsnePainting}, exhibit a reasonable connection to visual style by separating broad patterns, such as colourful paintings opposite black and white sketches, as well as smaller ones, such as unique styles of single artists. In order to correlate these embeddings with the collected socio-demographic information we must aggregate them to the artist-level. Consider the set of artists, A=$\{a^l|l=1,\dots,N_a\}$ where $N_a$ is the number of artists and each artist $a^l$ has a set of image embeddings $P^l=\{\textbf{e}^l_i|i=1,\dots,N^l\}$ where $N^l$ is the number of paintings for the $l$-th artist. For all further analysis we compute each artists centroid style embedding
\begin{equation}
    \textbf{c}^l=\frac{1}{N^l}\sum_i^{N^l}\textbf{e}^l_i.
\end{equation}
Only few artists have a singular repetitive style, which is especially true for art students for whom experimentation is essential. To be able to judge this variance of style we also compute the average intra-artist style distance to each centroid embedding with cosine distance $D_C$
\begin{equation}
    \sigma_{c}=\frac{1}{N_a}\sum_j^{N_a}\frac{1}{N^j}\sum^{N^j}_i D_{C}(\textbf{c}^l,\textbf{e}^j_{i})
\end{equation}
To have a comparable measure of variation we further compute the average centroid distance for all $N=\sum_i^{N_a} N^i$ images in the dataset
\begin{equation}
    \sigma_{c_{\text{global}}}=\frac{1}{N}\sum_j^{N_a}\sum^{N^j}_i D_{C}(\textbf{c}_N,\textbf{e}^j_{i}),
\end{equation}
where $\textbf{c}_N$ is the average of all image embeddings. The results are shown in Table~\ref{tab:distvar} for all three style embeddings. The \textit{Texture} embeddings have the smallest amount of variation, both globally and across artists. For the \textit{Archetype} embedding, the number of archetypes was set to $M=36$ through visual inspection of the reduction in the residual error as in~\cite{cutler1994archetypal}.

\begin{table}[t]
  \setlength{\tabcolsep}{7pt}
  \hspace{7pt}
    \begin{minipage}{.48\textwidth}
      \centering
        \caption{\textbf{Local and global style variation} with standard deviation.}
        \label{tab:distvar}
        \begin{tabular}{@{}lcc@{}}
        \toprule
         & $\sigma_{c}$  & $\sigma_{c_{\text{global}}}$ \\ \cmidrule(l){2-3} 
        \textit{VGG} & .283 $\pm$ .080 & .435 $\pm$ .101\\
        \textit{Texture} & .137 $\pm$ .049 & .211 $\pm$ .094\\
        \textit{Archetype} & .195 $\pm$ .121 & .323 $\pm$ .326\\ \bottomrule
        \end{tabular}
    \end{minipage}
    \hfill
    \begin{minipage}{.40\textwidth}
      \centering
        \caption{\textbf{Rank correlations of style and network distances.}}
        \label{tab:my-table}
        \setlength{\tabcolsep}{14pt}
        \begin{tabular}{@{}lcc@{}}
        \toprule
         & $\mathfrak{S}^{\text{U}}$ & $\mathfrak{S}^{\text{Y}}$ \\ \cmidrule(l){2-3} 
        \textit{VGG} &  .007 & -.032\\
        \textit{Texture} &  .043 & -.025   \\
        \textit{Archetype} & .012 & -.057\\ \bottomrule
        \end{tabular}
    \end{minipage}
    \hspace{12pt}
 \end{table}

\subsection{Social Networks and Style}

We use the node2vec algorithm~\cite{grover2016node2vec} on both graphs $\mathfrak{S}^{\text{U}}$ and $\mathfrak{S}^{\text{Y}}$ to project their relational data into a low-dimensional feature space. node2vec is a random-walk based graph embedding technique that preserves network neighbourhoods and, contrary to most other methods, structural similarity of nodes. This additional capability is especially useful for the larger network $\mathfrak{S}^{\text{Y}}$, in which the homophily captured by a pure social network such as $\mathfrak{S}^{\text{U}}$ is augmented by detailed and vast information on taste. We compute 128 node2vec features for each of the graphs and use cosine distance to generate a matrix of artist-level social network distances. Similarly, we generate pairwise style distances with the centroid embeddings $\textbf{c}^l$ for all three embeddings.

Spearmans’s rank coefficient is used to compute the correlation between the flattened upper triangular parts of the described distance matrices. The results in Table~\ref{tab:my-table} show that there are only very small correlations between stylistic and social distance. Even though the two graphs share only a minor similarity ($r_{sp}=.166$), neither network contains information that relates to inter-artist differences in style. The clear overlap between school affiliation and the smaller network graph $\mathfrak{S}^{\text{U}}$, as seen in Figure~\ref{fig:artistNetwork}, allows the further conclusion, that art schools too, have no bearing on artistic style.

\subsection{Socio-demographic Factors and Style}
\begin{figure}[t]
    \begin{subfigure}[t]{0.5\textwidth}
        \includegraphics[width=\textwidth]{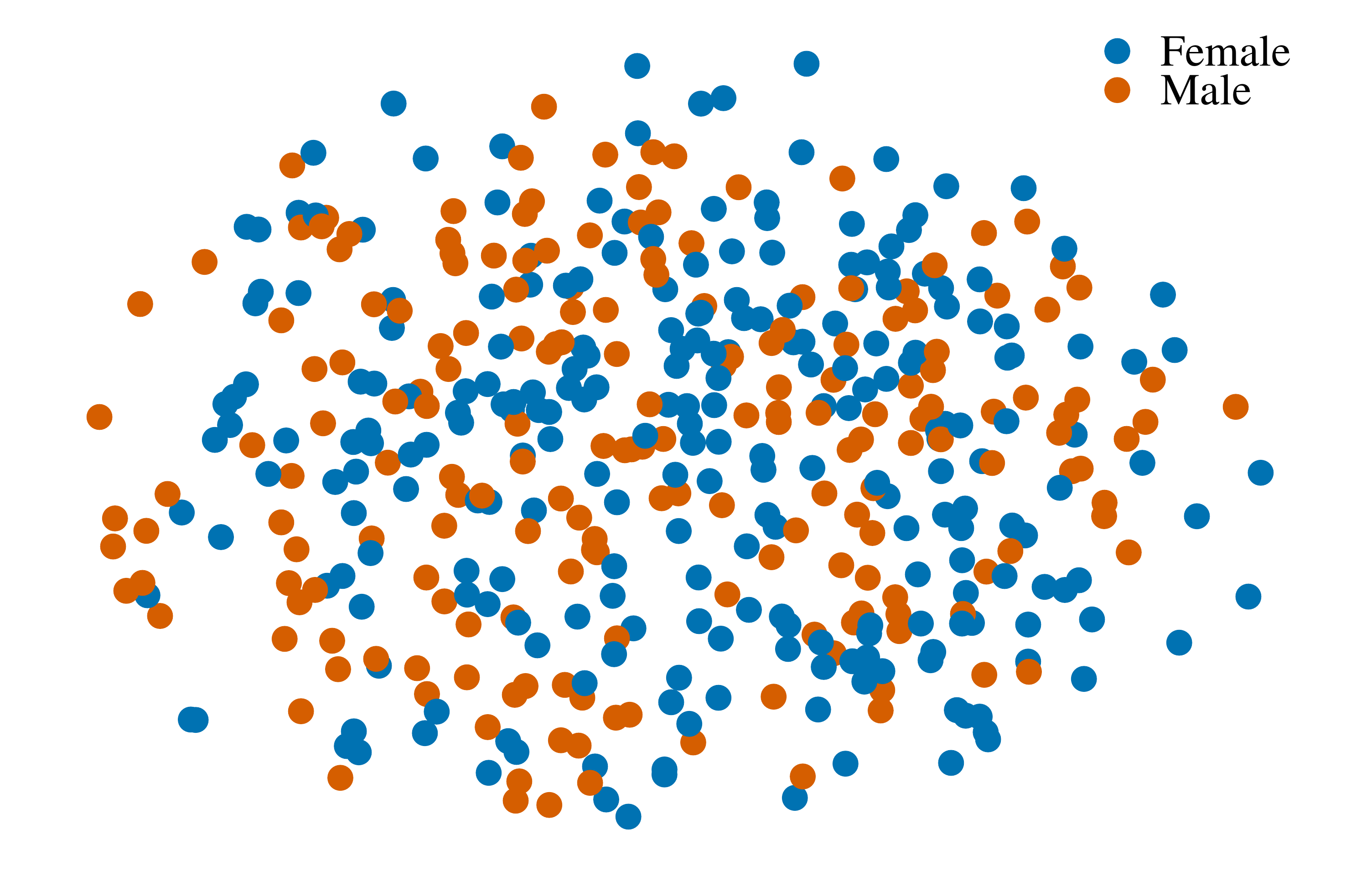}
    \end{subfigure}\hfill
    \begin{subfigure}[t]{0.5\textwidth}
        \includegraphics[width=\textwidth]{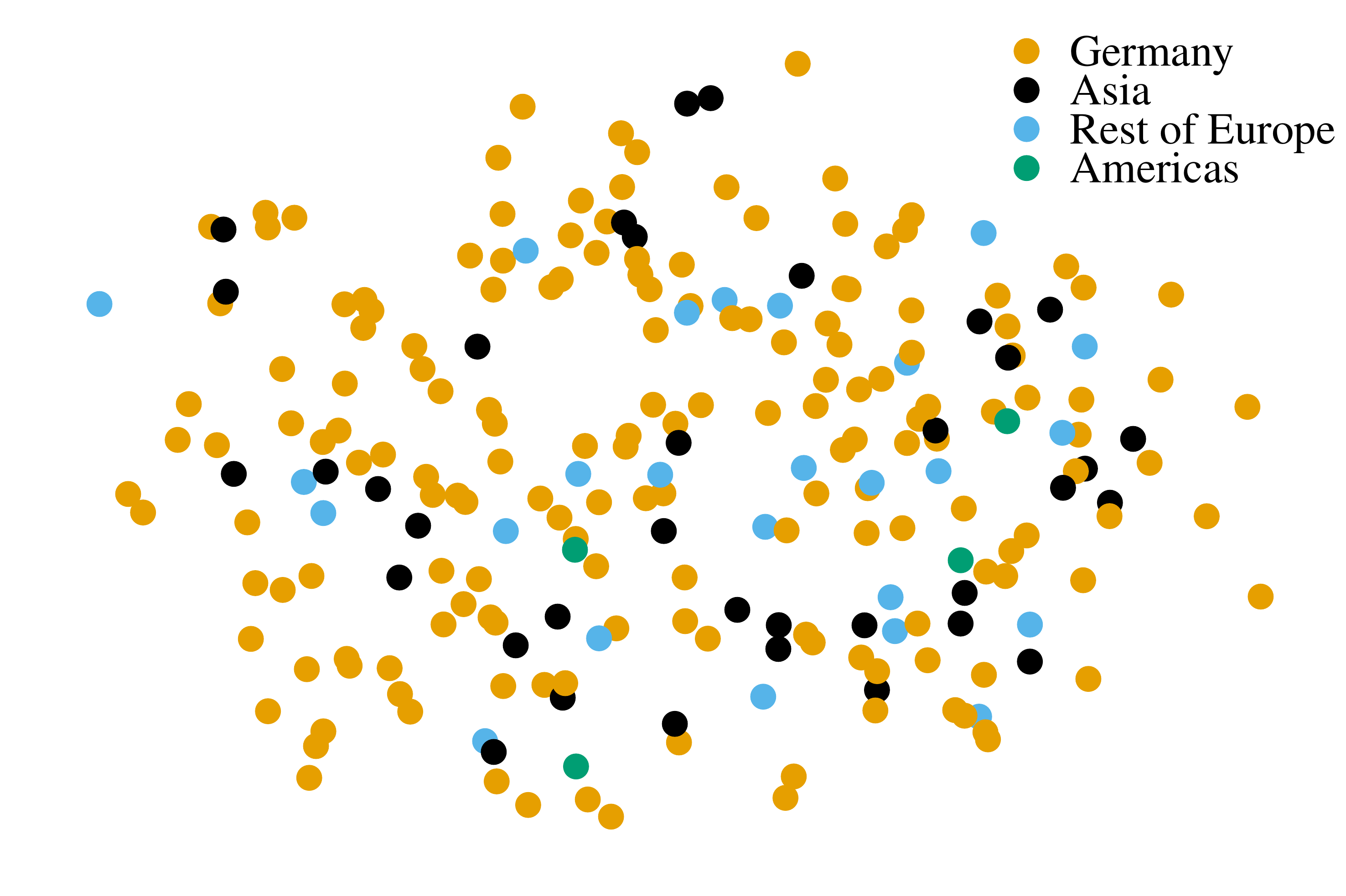}
    \end{subfigure}
    \caption{\textbf{The independence of artistic style and socio-demographic factors.} Visualisation of the centroid \textit{VGG} embeddings with t-SNE~\cite{maaten2008visualizing}.}
    \label{fig:datTsne}
\end{figure}

We investigate possible connections between the style embeddings and the collected data on the artists by jointly visualising them. Specifically, we extract a two-dimensional feature space from the \textit{VGG} embeddings with t-SNE~\cite{maaten2008visualizing}, both per image and per artist with the previously described aggregation. There were no visible patterns for any of the available variables, including Instagram-specific measures such as likes, comments or the number of followers and general ones such as nationality, gender or art school affiliation. We show two exemplary results in Figure~\ref{fig:datTsne}, in which the independence of style from these factors is apparent. This is not a surprising result as the non-visual factors are primarily attached to the individual artist and not their work. Even painting-specific reactions on Instagram depend more on the activity and reach of their creators than the artworks themselves. 

\section{Conclusion}

This work presented the first combined analysis of contemporary fine art and its social context by assembling a unique dataset on German art students and using unsupervised methodologies to extract and correlate artworks with their context. The collected data consisted of images, social network graphs and socio-demographic information on the artists. Three established methods to obtain style embeddings from images of paintings were briefly evaluated, outside of the usual framework of supervision, in their connection to common style annotations and general visual similarity. These embeddings of artistic style were shown to be entirely independent of any non-visual data. Further work will go into increasing dataset size, to reduce the effect of noise induced by the high amount of heterogeneity present in art produced by artists early in their career, and into contrasting the contemporary artworks with historical ones. 

\noindent
\textbf{Acknowledgement} This work was supported by JSPS KAKENHI No.~20K19822.

\clearpage

\bibliographystyle{eccv2020kit/splncs04}
\bibliography{eccv2020kit/eccv}

\end{document}